\documentclass[10pt,twocolumn,letterpaper]{article}
\usepackage{iccv}
\usepackage{times}
\usepackage{epsfig}
\usepackage{graphicx}
\usepackage{amsmath}
\usepackage{amssymb}
\usepackage{subcaption}
\usepackage{soul}
\usepackage{booktabs}
\usepackage{multirow}
\usepackage{tablefootnote}
\usepackage[pagebackref=true,breaklinks=true,letterpaper=true,colorlinks,bookmarks=false]{hyperref}

\iccvfinalcopy % *** Uncomment this line for the final submission

\def\iccvPaperID{***} % *** Enter the ICCV Paper ID here

\begin{document}
% \color[rgb]{1,0,0}
%%%%%%%%% TITLE
\title{Error-Aware Spatial Ensembles for Video Frame Interpolation}

% \and

\author{Zhixiang Chi$^{1}$,~~~
Rasoul Mohammadi Nasiri$^{1}$,~~~
Zheng Liu$^{1}$,~~~
Yuanhao Yu$^{1}$,~~~\\
Juwei Lu$^{1}$,~~~
Jin Tang$^{1}$,~~~ 
Konstantinos N Plataniotis$^{2}$\\
[0.1cm]
${^1}$Noah’s Ark Lab, Huawei Technologies \quad ${^2}$University of Toronto, Canada \\
% {\tt\small \{\{zhixiang.chi, rasoul.nasiri, zheng.liu1, Yuanhao.yu, tangjin, juwei.lu\}@huawei.com, kostas@ece.utoronto.ca}\\
}

% \author{Zhixiang Chi\inst{1},
% Rasoul Mohammadi Nasiri\inst{1},
% Zheng Liu\inst{1},
% Juwei Lu\inst{1},
% Jin Tang\inst{1}
% Konstantinos N Plataniotis\inst{2}}

% \institute{$^{1}$Noah’s Ark Lab, Huawei Technologies \quad $^{2}$University of Toronto, Canada\\
% \email{\{zhixiang.chi, rasoul.nasiri, zheng.liu1, tangjin, juwei.lu\}@huawei.com, kostas@ece.utoronto.ca}
% }

\maketitle
%\ificcvfinal\thispagestyle{empty}\fi

%%%%%%%%% ABSTRACT
\begin{abstract}

Video frame interpolation~(VFI) algorithms have improved considerably in recent years due to unprecedented progress in both data-driven algorithms and their implementations.  Recent research has introduced advanced motion estimation or novel warping methods as the means to address challenging VFI scenarios. However, none of the published VFI works considers the spatially non-uniform characteristics of the interpolation error (IE). This work introduces such a solution. By closely examining the correlation between optical flow and IE, the paper proposes novel error prediction metrics that partition the middle frame into distinct regions corresponding to different IE levels. Building upon this IE-driven segmentation, and through the use of novel error-controlled loss functions, it introduces an ensemble of spatially adaptive interpolation units that progressively processes and integrates the segmented regions. This spatial ensemble results in an effective and computationally attractive VFI solution.  Extensive experimentation on popular video interpolation benchmarks indicates that the proposed solution outperforms the current state-of-the-art (SOTA) in applications of current interest.

\end{abstract}

%%%%%%%%% BODY TEXT
\section{Introduction}

With recent advances in both high frame rate display and camera technology, it is possible to capture and display details of fast and complex motions. Playing high frame rate videos on displays with a high refresh rate benefits from smoother motion and higher visual quality~\cite{nasiri2015perceptual, katsenou2018exploring, nasiri2018temporal}. When it comes to low frame rate video input, VFI can be used to provide the needed temporal resolution ~\cite{park2020bmbc, niklaus2020softmax,chi2020all}.

\begin{figure}[t]
    \centering
    \begin{subfigure}{\linewidth} % First row
    \centering
    \includegraphics[width =0.49\linewidth]{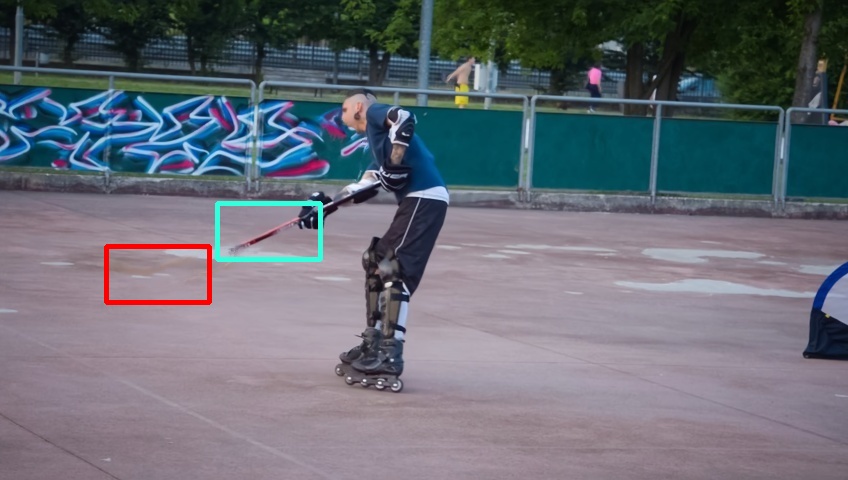}
    \includegraphics[width =0.49\linewidth]{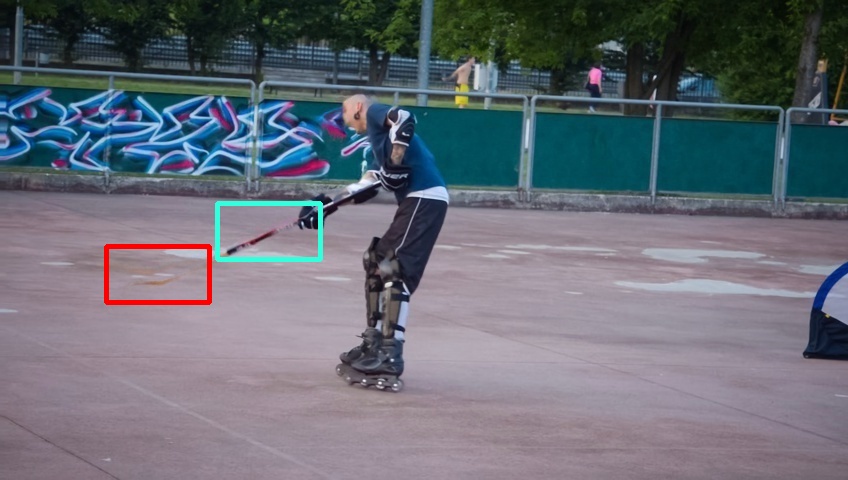}
    \end{subfigure}
    
    \vspace{0.03cm}
    \begin{subfigure}{0.49\linewidth}
    \centering
    \includegraphics[height=0.268\linewidth, width=0.49\linewidth]{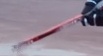}
    \includegraphics[height=0.268\linewidth, width=0.49\linewidth]{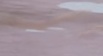}
    \caption{BMBC~\cite{park2020bmbc}}
    \end{subfigure}
    \begin{subfigure}{0.49\linewidth}
    \centering
    \includegraphics[height=0.268\linewidth, width=0.49\linewidth]{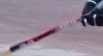}
    \includegraphics[height=0.268\linewidth, width=0.49\linewidth]{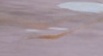}
    \caption{DAIN~\cite{DAIN}}
    \end{subfigure}
    \begin{subfigure}{\linewidth} % Second row
    \centering
    \includegraphics[width =0.49\linewidth]{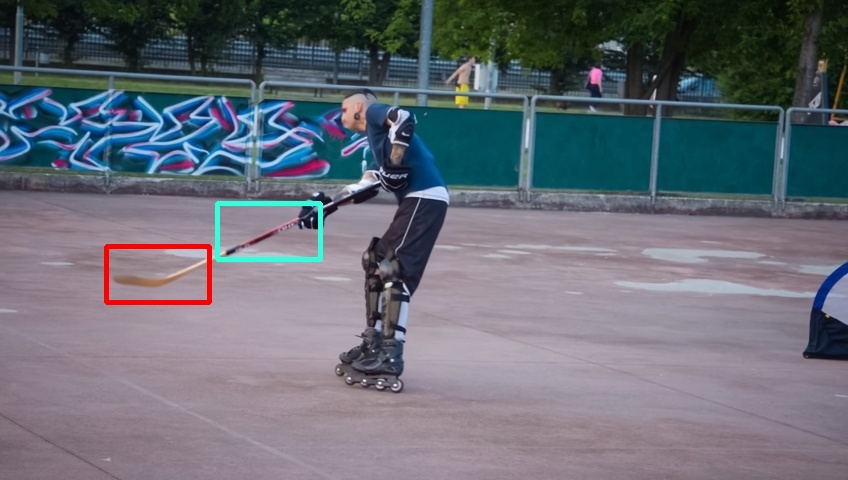}
    \includegraphics[width =0.49\linewidth]{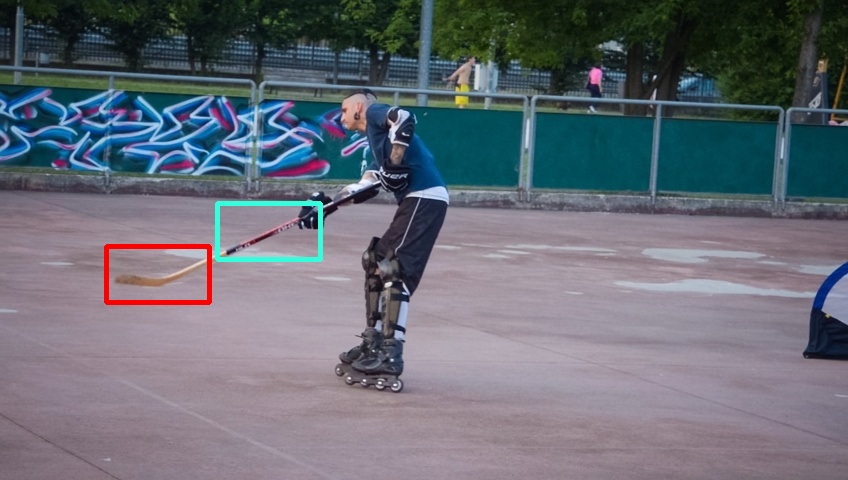}
    \end{subfigure}
    
    \vspace{0.03cm}
    \begin{subfigure}{0.49\linewidth}
    \centering
    \includegraphics[height=0.268\linewidth, width=0.49\linewidth]{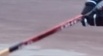}
    \includegraphics[height=0.268\linewidth, width=0.49\linewidth]{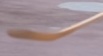}
    \caption{Ours~($\mathcal{L}_{ecc}$)}
    \end{subfigure}
    \begin{subfigure}{0.49\linewidth}
    \centering
    \includegraphics[height=0.268\linewidth, width=0.49\linewidth]{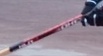}
    \includegraphics[height=0.268\linewidth, width=0.49\linewidth]{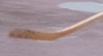}
    \caption{Ours~($\mathcal{L}_{ecp}$)}
    \end{subfigure}
    \caption{
    An example of frame interpolation. Our proposed error-aware method is more robust to handle the complex scenario (second example) where higher error is present.}
    \label{fig:front ims}
\end{figure}

Existing CNN-based VFI methods can be categorized as flow-based~\cite{park2020bmbc, niklaus2020softmax, liu2019deep, yuan2019zoom, niklaus2018context, jiang2018super, liu2017video, bao2018memc, DAIN, xu2019quadratic,chi2020all}, kernel-based~\cite{niklaus2017video_1, niklaus2017video} and phase-based~\cite{meyer2018phasenet}. Flow-based paradigm has been widely utilized and proven to be effective in benchmarking. These algorithms explicitly or implicitly handle complex motions by introducing different ideas such as higher order motion models~\cite{xu2019quadratic,chi2020all}, supplementary modules such as depth estimation, contextual feature extraction and blending masks estimation~\cite{niklaus2018context,DAIN,bao2018memc}. 

An observed shortcoming in the existing VFI methods is that despite the spatially non-uniform property of the interpolation error in frame, different regions of the middle frame are treated equally. Spatially adaptive processing has been successful in different applications but still missing in VFI for building more optimized solutions~\cite{li2019lap, suin2020spatially, zhang2019deep, lazebnik2006beyond}.

In VFI, interpolation errors are expected to be more centralized within or near the regions with complex motions. Thus, motion analysis represented by optical flow~(OF) is a guide to obtain and apply non-uniformity of IE. It should be noted that the widely popular flow-based methods never utilize any prior information conveyed by OF, such as spatially variant motion complexity, reflected by variation in motion vectors, although such information is readily available. Variations in motion vectors yields non-uniform interpolation complexities as well as non-uniformity of IE distributions across the frame. Obtaining IE distribution in regions of frame leads to more optimized processing of the frame. For example, in spatially uniform processing, the total loss is calculated by averaging across the whole frame while it can easily attenuate the optimization on such regions with higher error (complex motion)~\cite{yuan2019zoom}. In addition, the local level statistics exhibited by different regions can also be better explored~\cite{waleed2018unreasonable}. Therefore, an effective way is to separate the processing on the regions with distinguishable error levels to strengthen the optimization.

To this end, we dive deeper into the motion analysis and its relation with interpolation performance and propose an \textbf{E}rror-\textbf{A}ware adaptive framework for video \textbf{F}rame \textbf{I}nterpolation (EAFI). Concretely, we first estimate the OFs between two input video frames. We then propose our error prediction metrics based on OF to segment the frame into the regions according to IE levels. In addition, inspired by the Spatial Pyramid Matching~\cite{zhang2019deep, lazebnik2006beyond} and network ensemble methodologies~\cite{ensemble, lakshminarayanan2016simple}, we propose a spatial ensemble framework which progressively processes and assembles the segmented regions. Combined with our novel error controlled loss functions, we empirically show that the proposed ensemble approach with spatially adaptive processing and assembling is more effective. In contrast, we find that the uniform processing paradigm, can easily suffer from performance saturation~\cite{ensemble}. The proposed method outperforms the current SOTA on various benchmarks. See Fig.~\ref{fig:front ims} for a qualitative example compared with other alternatives. Furthermore, the error-adaptive framework allows us to design a compact model which performs comparably as DAIN~\cite{DAIN} but is \textbf{40} times smaller in model size.

This paper's contributions are manifold, as it: (1) proposes OF-based novel error prediction metrics; (2) introduces a heuristic to classify the interpolated regions as different levels of IE; (3) develops an error-aware, spatial ensemble framework to progressively process and assemble frame regions; (4) validates the solution using VFI benchmarks; (5) demonstrates SOTA performance, at a  superior speed-quality-size trade-off for a proposed compact model designed for low-power devices.

\section{Related work}

\noindent\textbf{Video frame interpolation:} 
Recent CNN-based methods that are specifically designed for VFI can be categorized as: kernel, phase, and flow-based. The kernel-based methods proposed by Niklaus \textit{et al.}~\cite{niklaus2017video_1,niklaus2017video} estimate the adaptive kernels for each pixel to weight its neighbor pixels. A phase-based method proposed by Meyer \textit{et al.}\cite{meyer2018phasenet} learns the phase decomposition of the middle frame. However, those approach are less accurate in dealing with complex motion.

Flow-based algorithms have shown reliable benchmark results by exploiting the motion information. Jiang \textit{et al.}~\cite{jiang2018super} estimated bidirectional OF along with a visibility map for VFI. Advanced warping methods are also developed to further improve the performance, such as adaptive warping layer~\cite{bao2018memc}, weighted warping according to depth information~\cite{DAIN}, and differential forward warping~\cite{niklaus2020softmax}.
Park \textit{et al.}~\cite{park2020bmbc} proposed bilateral motion estimation to address the holes and pixel overlapping from forward warping. Chi \textit{et al.}~\cite{chi2020all} and Xu \textit{et al.}~\cite{xu2019quadratic} proposed advanced higher-order motion models for accurate flow estimation. However, optical flow is an indispensable factor for flow-based methods; the information which it reveals about motion complexity has been ignored in designing VFI solutions. 

An effort towards this direction is by Yuan \textit{et al.}~\cite{yuan2019zoom} where they applied an instance-level adversarial training to strengthen the gradient and semantic information for small objects. However, a spatially adaptive method has not been fully explored to maximize the quality gain. 

\noindent\textbf{Spatially non-uniform processing: }Frame segmentation based on motion has been widely studied in video coding techniques, where spatially distinct areas are coded differently~\cite{dufaux1996segmentation,kunt1985second}. For other vision tasks, \cite{lazebnik2006beyond, zhang2019deep, hinton2015distilling} partitioned the images into equal-sized patches and then progressively processed and aggregated the information to achieve better performance. Li \textit{et al.}~\cite{li2019lap} proposed to process different regions in an image according to the difficulty levels. The regions are then integrated to achieve a better result.
% In contrast, our flow-based segmentation is measured by the error levels.

\noindent\textbf{Network ensembles: } Training multiple networks and averaging the predictions is known as Deep Ensembles~\cite{lakshminarayanan2016simple, ashukha2020pitfalls}. Under certain circumstances, it is more optimal compared to one single large network for classification~\cite{ensemble}. In this work, we propose a spatial ensemble framework to progressively integrate the segmented regions.

\begin{figure*}[t!]
    \centering
    \begin{subfigure}{\linewidth}
    \includegraphics[width =\linewidth, height=6cm]{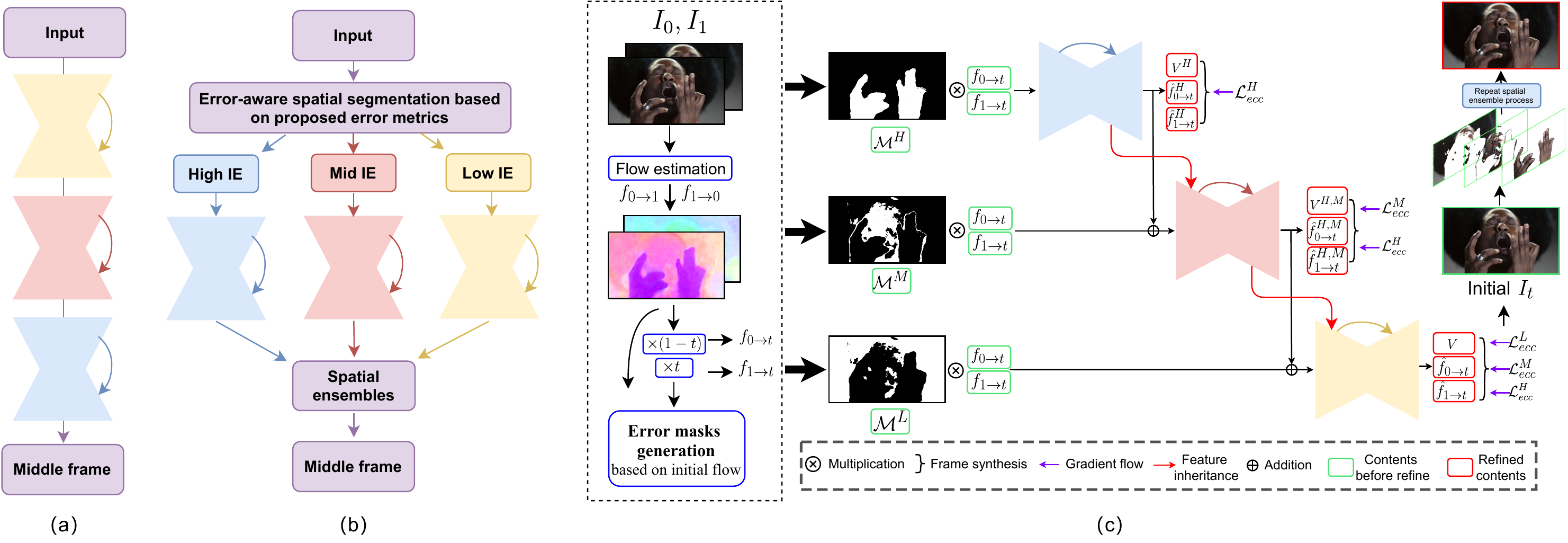}
    \end{subfigure}
    \caption{Comparison between different processing frameworks. (a) A single stacked network that uniformly processes the whole frame. (b) A naive spatially adaptive processing and assembling framework. (c) Proposed spatial ensemble framework, which progressively processes and assembles three regions in a frame and also the detailed drawing of error-aware segmentation used in proposed method and adaptive method (b).
    }
    \label{fig: architeture}
\end{figure*}

\section{Error-aware adaptive frame interpolation}

\subsection{Algorithm overview} 

An overview of the proposed method is illustrated in Fig.~\ref{fig: architeture}c. Given two consecutive video frames $I_0$ and $I_1$, we aim to generate the middle frame $I_t$ at $t \in (0,1)$. We first estimate the bidirectional OF $f_{0\rightarrow1}$ and $f_{1\rightarrow0}$ between $I_0$ and $I_1$ using the SOTA OF estimation network, RAFT~\cite{teed2020raft}. Then, the estimated OFs are scaled to obtain  $f_{0\rightarrow t}$ and $f_{1\rightarrow t}$ for the time step $t$. We then use our proposed OF-based error prediction metrics to partition the regions according to the predicted error levels. Finally, we progressively process different regions via our spatial ensemble framework to obtain the middle frame.

\subsection{Error-based frame segmentation using OF}

In nature, error in the interpolated frame is more likely accumulated in regions with large or variant motion. Therefore, it is more effective to adaptively process the regions to utilize the local semantic information and strengthen gradient flow~\cite{yuan2019zoom, zhang2019deep}. However, the underlying spatially variant motion information revealed by OF and the corresponding adaptive processing has been ignored~\cite{DAIN, bao2018memc, park2020bmbc, chi2020all}.

In this section, we propose three error prediction metrics based on statistical analysis on OF. Namely, motion size, motion variation, and photometric consistency. We finally merge the predicted error of these three metrics and make an error map. By thresholding the error map, three distinct error regions, high, mid, and low, are obtained, which are used in the adaptive processing in later step. In the following, we introduce the proposed error metrics using $f_{0\rightarrow 1}$ and $f_{0\rightarrow t}$. $f_{1\rightarrow 0}$ and $f_{1\rightarrow t}$ will be utilized in the same way to produce corresponding error metrics. For each metric, we take the maximum values between metrics yielded by OF from both directions as its final metric.

\noindent\textbf{Motion size:} In videos, large motion is caused by either fast moving objects or severe camera movement. Intuitively, it is counted as a source of IE as a possible range of matching for finding OF is increased. In other words, larger spatial search space for matching the pixels between consecutive frames increases uncertainties and the likelihood of errors.

The motion vector in OF gives the underlying information about which pixels or regions in a frame are moving at a large step size in a unit time step between two frames. So, we use the magnitude of OF in each pixel as a motion metric $ms$ to predict IE distribution:
\begin{equation}
    \mathcal{E}_{ms} = \left \| f_{0\rightarrow{t}} \right \|.
\end{equation}

\noindent\textbf{Motion variation:}
Uniform motion in a region of frame usually represents rigid object motion, which is relatively easy for interpolators. In contrast, some variations in motion between neighbor pixels represent small parts and non-rigid body motion, which is challenging. Therefore, we propose to estimate such error using variation in motion. This variation for a sample direction of $x$ is defined as:
\begin{equation}  
    \mathcal{E}^x_{mv} = \left \|f^x_{0 \rightarrow t} - \widetilde{f}^x_{0 \rightarrow t}\right \|_2,
\end{equation}
where $\widetilde{f}^x$ denotes the mean value of the OF in $x$ direction. 
For simplicity, we measure this metric for two directions of $x$ and $y$ as $\mathcal{E}^x_{mv}$ and $\mathcal{E}^y_{mv}$, separately. The final motion variation metric $\mathcal{E}_{mv}$ is calculated as the maximum value between $\mathcal{E}^x_{mv}$ and $\mathcal{E}^y_{mv}$. Motion variation is a statistical metric to estimate error from local variations of motion. This could happen even in small motions, which makes it somewhat independent of $\mathcal{E}_{ms}$. 

\noindent\textbf{Photometric consistency:} $\mathcal{E}_{ms}$ and $\mathcal{E}_{mv}$ are directly defined on OF maps. They target common cases where high IE can potentially happen. In addition, we propose another metric, which is defined at the pixel level. Specifically, we apply $f_{0 \rightarrow 1}$ to warp $I_1$ and calculate IE within the warped $I_1$ as an estimation of IE for $I_t$. 
We define this metric as a photometric consistency, $\mathcal{E}_{pc}$, as follows:
\begin{equation}
    \mathcal{E}_{pc} = \left \| I_0 - w(I_1, f_{0 \rightarrow 1}) \right \|_1,
\end{equation}
where $w(I, f)$ is the backward warping function and $\left \|  \cdot \right \|_1$ is the $l_1$ norm. Photometric consistency is also used as a metric for occlusion detection~\cite{niklaus2020softmax, baker2011database}. 

\begin{figure*}
    \centering
    \begin{subfigure}{\linewidth}
    \includegraphics[width =\linewidth]{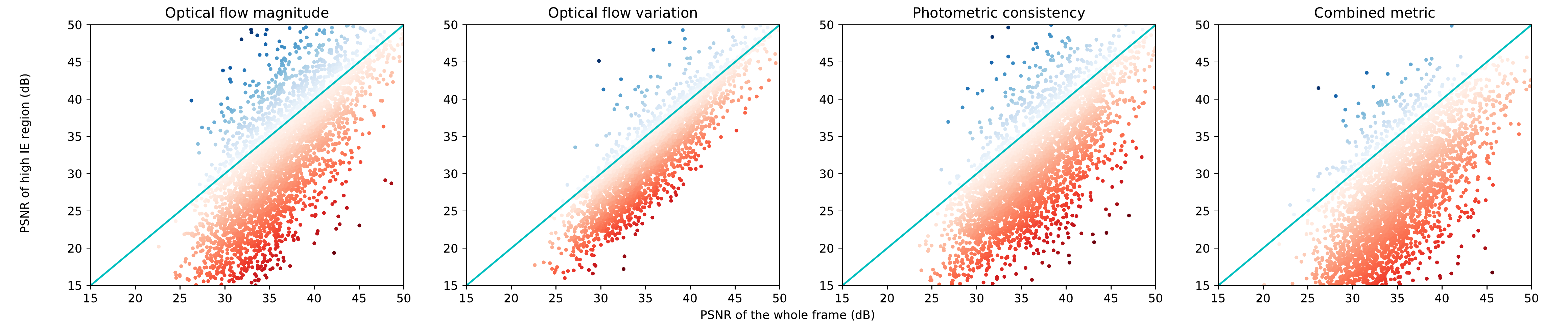}
    \end{subfigure}
    \caption{PSNR comparison between the high IE regions and whole frame classified by four proposed error prediction metrics $\mathcal{E}_{ms}, \mathcal{E}_{mc}$, $ \mathcal{E}_{pc}$ and $ \mathcal{E}_{tot}$. The evaluation is performed on validation set and a model without adaptive processing.}
    \label{fig: PSNR scatter}
\end{figure*}

\noindent\textbf{Error metric: } The three proposed error prediction metrics convey different clues in generating the error map. To maximize the accuracy in error prediction, we propose to integrate all metrics. We first normalize $\mathcal{E}_{ms}$, $\mathcal{E}_{mv}$, and  $\mathcal{E}_{pc}$ and then merge them as:
\begin{equation}
    \mathcal{E}_{tot} = max(\frac{\mathcal{E}_{ms}}{\Gamma_{ms}},
                        \frac{\mathcal{E}_{mv}}{\Gamma_{mv}},
                        \frac{\mathcal{E}_{cp}}{\Gamma_{cp}}),
\end{equation}
where ${\Gamma_{ms}}$, ${\Gamma_{mv}}$, and ${\Gamma_{cp}}$ are the maximum values of $\mathcal{E}_{ms}, \mathcal{E}_{mv}$ and $\mathcal{E}_{cp}$ in each frame, respectively. The normalized estimated error map $\mathcal{E}_{tot}$ is then thresholded to obtain three binary error masks $\mathcal{M}^H$, $\mathcal{M}^M$, and $\mathcal{M}^L$ for high, mid, and low level error regions for each frame as:
\begin{equation}
% \label{threshold eq}
\begin{cases}
    \centering
    \mathcal{M}^{H} = 1 & \text{if $\mathcal{E}_{tot}$ $\geq \tau^H_\mathcal{E}$}\\
    \mathcal{M}^M = 1 & \text{if $\tau^H_\mathcal{E}>$ $\mathcal{E}_{tot}$  $\geq \tau^M_\mathcal{E}$}\\
    \mathcal{M}^L = 1 & \text{if $\tau^M_\mathcal{E}>$ $\mathcal{E}_{tot}$}.
  \end{cases}
  \label{three_thresh}
\end{equation}
$\tau_\mathcal{E}$ denotes the thresholds. We set $\tau^M_\mathcal{E}$ as the mean of $\mathcal{E}_{tot}$ and set $\tau^H_\mathcal{E}$ as $((1-\tau^M_\mathcal{E})/2 + \tau^M_\mathcal{E})$ to equally divide the higher error interval. Fig.~\ref{fig: PSNR scatter} shows the PSNR comparison between high IE regions and the whole frame. As more points are dispersedly distributed below the equal line, it indicates that the segmentation based on the proposed metrics is meaningful.

\subsection{Error-aware adaptive processing}\label{adaptive}
\noindent\textbf{Adaptive spatial ensemble framework: }
The OFs to the middle frame, $f_{0 \rightarrow t}$ and $f_{1 \rightarrow t}$, are directly estimated from $f_{0 \rightarrow 1}$ and $f_{1 \rightarrow 0}$. It may generate artifacts near the boundaries~\cite{jiang2018super} as the flow estimation is not tailored for frame interpolation~\cite{xue2019video}. Following~\cite{jiang2018super, chi2020all}, we propose a frame synthesis module to refine the flows and generate the target middle frame. We observed that processing the whole frame uniformly using a single network (Fig.~\ref{fig: architeture}a) is not optimal, and simply increasing the network size does not improve much~\cite{ensemble, zhang2019deep}. Benefited from error-aware region partitioning, we propose to process the regions adaptively so that each region can better focus on its local information. On the other hand, the loss calculation will not be disturbed by other regions. Inspired by~\cite{lakshminarayanan2016simple}, a naive way is to apply independent processing on each of the three regions followed by regional integration as shown in Fig.~\ref{fig: architeture}b. However, we find that, with proper information sharing among processing units, the network achieves better performance~\cite{zhang2019deep}.

We propose an alternative of Fig.~\ref{fig: architeture}b which strongly bonds the processing in different regions, as shown in Fig.~\ref{fig: architeture}c. Our framework is inspired by~\cite{zhang2019deep}, where fine-to-coarse processing is utilized. The network processes three regions one by one in stages of Fig.~\ref{fig: architeture}c, starting from high IE at the top stage and then mid and low IE in the next stages. The refined OF at the output of each stage is then added to the region waiting to be refined at the subsequent stage. It allows the higher IE regions to be processed with more layers and helps the processing of other regions. Also, the feature maps in the decoder of the refined stage are passed to the encoder of the next stage. By inheriting more abstract features from the refined regions, the subsequent stages can avoid computing them redundantly. Thus, each stage is specialized in processing its assigned regions for OF, and in the meantime, provides auxiliary error reduction (improvement) on the regions already refined by the former stage. 

For each stage, we employ residual learning to learn the OF residuals. It also generates a visibility map $V$; thus, each region can be synthesized as in ~\cite{jiang2018super, chi2020all}:
\begin{equation}
    I_t^r= V^r\odot g(I_0, \hat{f}^r_{0\rightarrow{t}}) + (1-V^r)\odot g(I_1,  \hat{f}^r_{1\rightarrow{t}}),
\end{equation}
where $g(I, f)$ is the forward warping function~\cite{niklaus2020softmax}, which takes a flow map $f$ to warp an image $I$, and $\odot$ is the element-wise multiplication. $\hat{f}$ denotes the refined OF, and $r$ is the choice of regions $\{H, M, L \}$. We adopt U-Net~\cite{ronneberger2015u} architectures for all processing units. The detailed architecture of our framework is provided in the supplement.

% compare SOTA
\begin{table*}[t]
% \scriptsize
% \small
\footnotesize
\centering
\setlength{\tabcolsep}{2.8pt} % Default value: 6pt

\begin{tabular}{lccccccccccccc}
\toprule
\multirow{2}{*}{Methods} & Training & \#Param & \multicolumn{3}{c}{Vimeo90K~\cite{xue2019video}} & \multicolumn{3}{c}{UCF101(w/o mask)~\cite{liu2017video}} &
\multicolumn{2}{c}{UCF101(w/ mask)} & \multicolumn{3}{c}{Middlebury~\cite{baker2011database}} \\
 \cmidrule(r){4-6} \cmidrule(r){7-9} \cmidrule(r){10-11} \cmidrule(r){12-14} 
 & data & (million) & PSNR $\uparrow$ & SSIM $\uparrow$ & LPIPS $\downarrow$ & PSNR $\uparrow$ & SSIM $\uparrow$ & LPIPS $\downarrow$ & PSNR $\uparrow$ & SSIM $\uparrow$ & PSNR $\uparrow$ & SSIM $\uparrow$ & LPIPS $\downarrow$\\
 \midrule
 DVF & Vimeo90K & 4.72 & 32.64 & 0.950 & 0.031 & 34.11 & 0.941 & 0.033 & 29.37 & 0.861 & 29.56 & 0.860 & 0.100\\
 ToFlow & Vimeo90K  & \textcolor{cyan}{2.70} & 33.73 & 0.952 & 0.027 & 34.58 & 0.947 & 0.027 & 30.09 & 0.877 & 35.29 & 0.956 & 0.024 \\
 SepCov-$_{L1}$ & Proprietary & 21.6 & 33.80 & 0.956 & 0.027 & 34.79 & 0.947 & 0.029 & 30.03 & 0.869 & 35.73 & 0.959 & 0.017 \\
 SuperSloMo & Vimeo90K & 50.6 & 34.35 & 0.957 & 0.022 & 34.75 & 0.947 & 0.025 & 30.22  & 0.880 & 36.76 & 0.964 & 0.019 \\
 MEMC-Net & Vimeo90K& 70.3 & 34.29 & 0.962 & 0.027 & 35.01 & \textcolor{cyan}{0.951} & 0.030 & 30.34 & 0.881 & 36.48 & 0.965 &  0.020\\
 DAIN & Vimeo90K & 24.0 & 34.70 & 0.964 & 0.022 & 35.00 & 0.950 & 0.028 & 30.31 &  0.879& 36.70 & 0.965 & 0.017 \\
 BMBC & Vimeo90K & 11.0 & 35.01 & 0.964 & 0.023 & 35.15 & 0.950 & 0.029 & 30.54 & 0.884 & 36.77 & 0.965 & 0.021\\
 SoftSplat-$_{L_{Lap}}$ & Vimeo90K & 7.80 & \textcolor{cyan}{36.10} & \textcolor{cyan}{0.970} & 0.021 & \textcolor{cyan}{35.39} & \textcolor{red}{0.952} & 0.033 & \textcolor{red}{30.80} & \textcolor{cyan}{0.888} & \textcolor{cyan}{38.42} & \textcolor{cyan}{0.971} & \textcolor{cyan}{0.016} \\
 SoftSplat-$_{L_{F}}$ & Vimeo90K & 7.80 & 35.48 & 0.964 & \textcolor{cyan}{0.013} & 35.10 & 0.948 & \textcolor{cyan}{0.022} & 30.51 & 0.880 & 37.55 & 0.965 & \textcolor{red}{0.008}\\
 \midrule
 Ours-compact & Vimeo90K& \textcolor{red}{0.59} & 35.01 & 0.963 & 0.023 & 35.13 & 0.950 & 0.029 & 30.48 & 0.883 & 36.91 & 0.965 &  0.018\\
 Ours-$\mathcal{L}_{ecp}$ & Vimeo90K& 12.4 & 35.95 & 0.967 & \textcolor{red}{0.012} & 35.24 & 0.949 & \textcolor{red}{0.020} & \textcolor{cyan}{30.62} & 0.883 & 37.65 & 0.967 & \textcolor{red}{0.008}  \\
 Ours-$\mathcal{L}_{ecc}$ & Vimeo90K& 12.4 & \textcolor{red}{36.38} & \textcolor{red}{0.972} & 0.020 & \textcolor{red}{35.41} & \textcolor{red}{0.952} & 0.031 & \textcolor{red}{30.80} & \textcolor{red}{0.889} & \textcolor{red}{38.70} & \textcolor{red}{0.973} & 0.017 \\

\bottomrule
\end{tabular}
\caption{Performance comparison with the state-of-the-art methods on widely used datasets. The numbers in \textcolor{red}{red} and \textcolor{cyan}{cyan} indicate the first and second best results respectively. Most results of the compared methods are copied from~\cite{niklaus2020softmax}.
}
\label{tab: compare SOTA}
% \vspace{-3mm}
\end{table*}

\noindent\textbf{Adaptive post processing: }
The generated middle frame in the OF-based frame synthesis may still contain artifacts. Situations such as occlusion or changes in color raise the need to exceed motion-based synthesis. Thus, we repeat the same spatial ensemble framework for pixel-level refinement as post-processing. At this step, since the refined OF, $\hat{f}_{0 \rightarrow t}$ and $\hat{f}_{1 \rightarrow t}$, is more tailored for frame interpolation, the error masks can be updated. Furthermore, we input $I_0$ and $I_1$ and their warped version to all stages to compensate for the information loss during synthesis.

% \subsection{Error controlled loss functions}\label{loss}
\noindent\textbf{Error controlled loss functions: }
Due to the non-uniform spatial distribution of IE, optimizing the loss over the whole frame impairs the quality in the higher IE regions~\cite{shen2018deep,yuan2019zoom}. We contribute our three error masks to propose error controlled content loss for the proposed ensembles: 
\begin{equation}
\small
\mathcal{L}_{ecc}= \sum_{r \in H,M,L}\mathcal{L}^r_{ecc} = \sum_{j=1}^{3} \sum_{r \in H,M,L}\frac{\left \| \mathcal{M}^r\odot (I_t^j - {I_{gt}}) \right \|_1 }{\left \| \mathcal{M}^r \right \|_1},
\label{eq: ecc}
\end{equation}
where, $\mathcal{M}^*$ are the error masks, $I_{gt}$ is the ground truth, $I_t^j$ is the synthesized region at stage $j$. We also adopt a feature-based perceptual loss to address the blurriness potentially happens in rich textures and motion boundaries~\cite{niklaus2018context,niklaus2017video}. We follow~\cite{niklaus2018context} to apply an error controlled perceptual loss:
\begin{equation}
\small
\mathcal{L}_{ecp}=\sum_{j=1}^{3} \sum_{r \in H,M,L}\left \| \mathcal{M}^r\odot \phi (I_t^j) - \mathcal{M}^r\odot{}\phi (I_{gt}) \right \|_2,
\label{eq: ecp}
\end{equation}
where $\phi(\cdot)$ are the feature maps generated at conv$4\_4$ layer of a pre-trained VGG19 network~\cite{simonyan2014very}. Note, for the first and second stages, the loss terms are omitted when the masks are not present. The error-controlled losses strongly constrain the regions in ${I_t}$ as each region exhibits its own local semantic details. Therefore, each stage is enforced to be specialized in the process of each region; thus, better optimization is achieved. To compute the loss for post-processing module, we simply update the error masks and replace $I_t^j$ with the refined region $\hat{I}_t^j$ in (\ref{eq: ecc}) and (\ref{eq: ecp}).

% compare middleburry
\begin{table*}[t]
\setlength{\tabcolsep}{5pt} % Default value: 6pt
\renewcommand{\arraystretch}{1} % Default value: 1
\footnotesize
\centering

\begin{tabular}{lccccccccccccccccc}
\toprule
& Training & \multicolumn{2}{c}{Mequon} & \multicolumn{2}{c}{Schefflera	
} & \multicolumn{2}{c}{Urban	
} & \multicolumn{2}{c}{Teddy	
} & \multicolumn{2}{c}{Backyard
} & \multicolumn{2}{c}{Basketball	
} & \multicolumn{2}{c}{Dumptruck	
} & \multicolumn{2}{c}{Evergreen	
}\\

\cmidrule(r){3-4} \cmidrule(r){5-6} \cmidrule(r){7-8} \cmidrule(r){9-10} \cmidrule(r){11-12} \cmidrule(r){13-14} \cmidrule(r){15-16} \cmidrule(r){17-18} 

& data & IE & NIE & IE & NIE & IE & NIE & IE & NIE & IE & NIE & IE & NIE & IE & NIE & IE & NIE\\ 
\midrule
SuperSlomo& Proprietary &2.51&	0.59&	3.66&	0.72&	2.91&	0.74&	5.05&	0.98&	9.56&	0.94&	5.37&	0.96&	6.69&	0.6&	6.73&	0.69 \\
SepCov-$\mathcal{L}1$& Proprietary &	2.52&	0.54&	3.56&	0.67&	4.17&	1.07&	5.41&	1.03&	10.2&	0.99&	5.47&	0.96&	6.88&	0.68&	6.63&	0.70\\
DAIN& Vimeo90K &2.38&	0.58&	3.28&	0.60&	3.32&	0.69&	4.65&	0.86&	\textcolor{cyan}{7.88}&	\textcolor{cyan}{0.87}&	4.73&	0.85&	6.36&	0.59&	6.25&	0.66 \\
BMBC& Vimeo90K&	2.30&	0.57&	3.07&	0.58&	3.17&	0.77&	4.24&	0.84&	\textcolor{red}{7.79}&	\textcolor{red}{0.85}&	\textcolor{red}{4.08}&	\textcolor{cyan}{0.82}&	\textcolor{cyan}{5.63}&	\textcolor{cyan}{0.58}&	\textcolor{cyan}{5.55}&	\textcolor{red}{0.56}\\
SoftSplat& Vimeo90K & \textcolor{red}{2.06} &	\textcolor{cyan}{0.53}&	\textcolor{cyan}{2.80}& \textcolor{cyan}{0.52}&	\textcolor{cyan}{1.99}&	\textcolor{cyan}{0.52}&	\textcolor{cyan}{3.84}&	\textcolor{cyan}{0.80}&	8.10&	\textcolor{red}{0.85}&	\textcolor{cyan}{4.10}&	\textcolor{red}{0.81}&	\textcolor{red}{5.49}&	\textcolor{red}{0.56}&	\textcolor{red}{5.40}&	\textcolor{cyan}{0.57}\\
Ours& Vimeo90K&	\textcolor{cyan}{2.10}&	\textcolor{red}{0.50}&	\textcolor{red}{2.54}&	\textcolor{red}{0.46}&	\textcolor{red}{1.77}&	\textcolor{red}{0.42}&	\textcolor{red}{3.82}&	\textcolor{red}{0.79}&	9.04&	\textcolor{red}{0.85}&	4.80&	\textcolor{red}{0.81}&	5.89&	\textcolor{cyan}{0.58}&	5.77&   \textcolor{cyan}{0.57}\\
\bottomrule
\end{tabular}
\caption{Evaluation on the Middlebury benchmark for both Interpolation Error~(IE) and Normalized Interpolation Error~(NIE). %We report quantitative results in terms of both Interpolation Error~(IE) and Normalized Interpolation Error~(NIE).
}
\label{tab: compare SOTA middle}
% \vspace{-4mm}
\end{table*}

\section{Experiments}

\noindent\textbf{Training: } We perform training on Vimeo90K  dataset~\cite{xue2019video}. It contains 51,312 samples of frame triplets with a resolution of 256$\times$448. We randomly choose 3500 sequences for validation. For data augmentation, we randomly flip the frames horizontally and vertically as well as reverse the temporal order of the whole sequence. We train the network to interpolate the middle frame~($t=0.5$) using the Adam optimizer~\cite{kingma2014adam}. We adopt the stage-wise training strategy as in~\cite{zhang2018densely}. 
The processing unit for high IE regions is trained first, followed by mid and low IE regions separately with a learning rate of $10^{-4}$ for 20 epochs. The entire network is then jointly trained with a learning rate of $10^{-5}$. The model converges after 120 epochs with a batch size of 6. All the experiments are conducted using Nvidia P100 GPUs.

\noindent\textbf{Evaluation datasets and metrics:}  We perform the evaluation on several well-known datasets: UCF101~\cite{liu2017video} (379 triplets), Vimeo90K~\cite{xue2019video} (3782 triplets), Middlebury~\cite{baker2011database} (12 and 8 sequences). We also conduct evaluations on GoPro~\cite{nah2017deep} and DAVIS~\cite{pont20172017} datasets, which are initially designed for deblurring and segmentation tasks and are commonly used in benchmarking interpolation solutions. We extract 1355 sequences of 9 frames from GoPro~\cite{nah2017deep} for multi-frame interpolation at arbitrary time steps. We also extract 2068 triplets from DAVIS~\cite{pont20172017} to evaluate challenging large motions. For UCF101, we also evaluate the quality of high motion regions by applying masks provided by~\cite{liu2017video}. The code provided by~\cite{jiang2018super} and~\cite{niklaus2020softmax} is used for evaluation. As for performance metrics, we compute PSNR, SSIM to measure the pixel-level similarities and also adopt LPIPS~\cite{zhang2018unreasonable} to measure the perceptual quality of the interpolated frames. 

\subsection{Comparison with the state-of-the-art}\label{compare sota}

% Comparison on GoPro and DAVIS
\begin{table}
% \scriptsize
\small
% \footnotesize
\centering
\setlength{\tabcolsep}{3.2pt} % Default value: 6pt

\begin{tabular}{lcccccc}
\toprule
\multirow{2}{*}{Methods} & \multicolumn{3}{c}{GoPro~\cite{nah2017deep} (x8)} & \multicolumn{3}{c}{DAVIS~\cite{pont20172017} (x2)}\\
 \cmidrule(r){2-4} \cmidrule(r){5-7}
 & PSNR & SSIM & LPIPS & PSNR & SSIM & LPIPS \\
 & $\uparrow$ & $\uparrow$ & $\downarrow$ & $\uparrow$ & $\uparrow$ & $\downarrow$ \\
\midrule
SepConv-$\mathcal{L}1$ & 28.96& 0.869& 0.072&25.47 & 0.753& 0.118\\
BMBC  &29.22 &0.874 &0.058 &26.49 & 0.791&0.121 \\
DAIN  & 29.21&0.876 &0.055 & 27.20&0.814 &  0.083\\
\midrule
Ours-$\mathcal{L}_{ecp}$  &29.47 &0.874 &\textbf{0.043} &26.98 & 0.792& \textbf{0.080}\\
Ours-$\mathcal{L}_{ecc}$  &\textbf{29.79} &\textbf{0.882} &0.059 & \textbf{27.55} & \textbf{0.815}& 0.111\\

\bottomrule
\end{tabular}
\caption{Comparison on GoPro and DAVIS datasets.
}
\label{tab: Gopro&davis}
% \vspace{-3mm}
\end{table}

\noindent\textbf{Quantitative analysis:} We compare the proposed method with several state-of-the-art frame interpolation methods: DVF \cite{liu2017video}, ToFlow~\cite{xue2019video}, SepConv-\textit{L}$_1$~\cite{niklaus2017video}, MEMC-Net~\cite{bao2018memc}, DAIN~\cite{DAIN}, SuperSloMo~\cite{jiang2018super}, SoftSplat~\cite{niklaus2020softmax} and BMBC~\cite{park2020bmbc}. All the methods are trained on Vimeo90K except SepConv, thus, the comparison is fair and faithful. The quantitative results reported in Table.~\ref{tab: compare SOTA} show that the proposed method consistently performs favorably against the existing approaches on all three benchmarks. 
% Note that, we did not add QVI~\cite{xu2019quadratic} to this table as it requires different benchmark settings other than official triplets. And for fair comparison, we train our method and QVI on a same public dataset. Due to the page limitation, we include the detailed comparison in the supplement. 
We also submitted our results of the Middlebury Evaluation set to the benchmark server. Our error-aware approach currently ranks the $\textbf{1}^\textbf{st}$ and $\textbf{4}^\textbf{th}$ in terms of NIE and IE, respectively. Table.~\ref{tab: compare SOTA middle} shows the detailed comparison on each sequence; our methods generate favorable results against the compared methods.

Table.~\ref{tab: Gopro&davis} reports the performance comparison on DAVIS dataset~\cite{pont20172017}. Our methods perform better when handling challenging complex motions. In addition, the proposed methods are also able to interpolate at any time step $t \in (0, 1)$. Thus, we evaluate the performance of multi-frame interpolation by interpolating 7 frames at $t_i=\frac{i}{8}, i \in [1, 2, ..., 7]$ on GoPro~\cite{nah2017deep}. Note that the frame resolution of GoPro~\cite{nah2017deep} is 720$\times$1280, which further challenges interpolators as motion becomes much larger. Despite that, our methods still outperform others by a large margin, as shown in Table.~\ref{tab: Gopro&davis}. One interesting observation is that the performance of VFI methods is dependent to the frame rate of the target videos in training data. Vimeo90k target videos are 30fps, while GoPro is designed for 240fps videos. We performed additional investigation in supplement by involving QVI~\cite{xu2019quadratic} method with original target videos of 960fps.
% \hl{An interesting thing we observed is that, depending on the target video frame rate, the training videos are important. Vimeo90K are extracted from 30fps videos, whereas GoPro has 240fps videos. We perform an additional study to investigate the impact of training video frame rate along with }QVI~\cite{xu2019quadratic} \hl{in the supplement.}

\begin{figure}
    \centering
    \begin{subfigure}{\linewidth}
    \includegraphics[width =\linewidth]{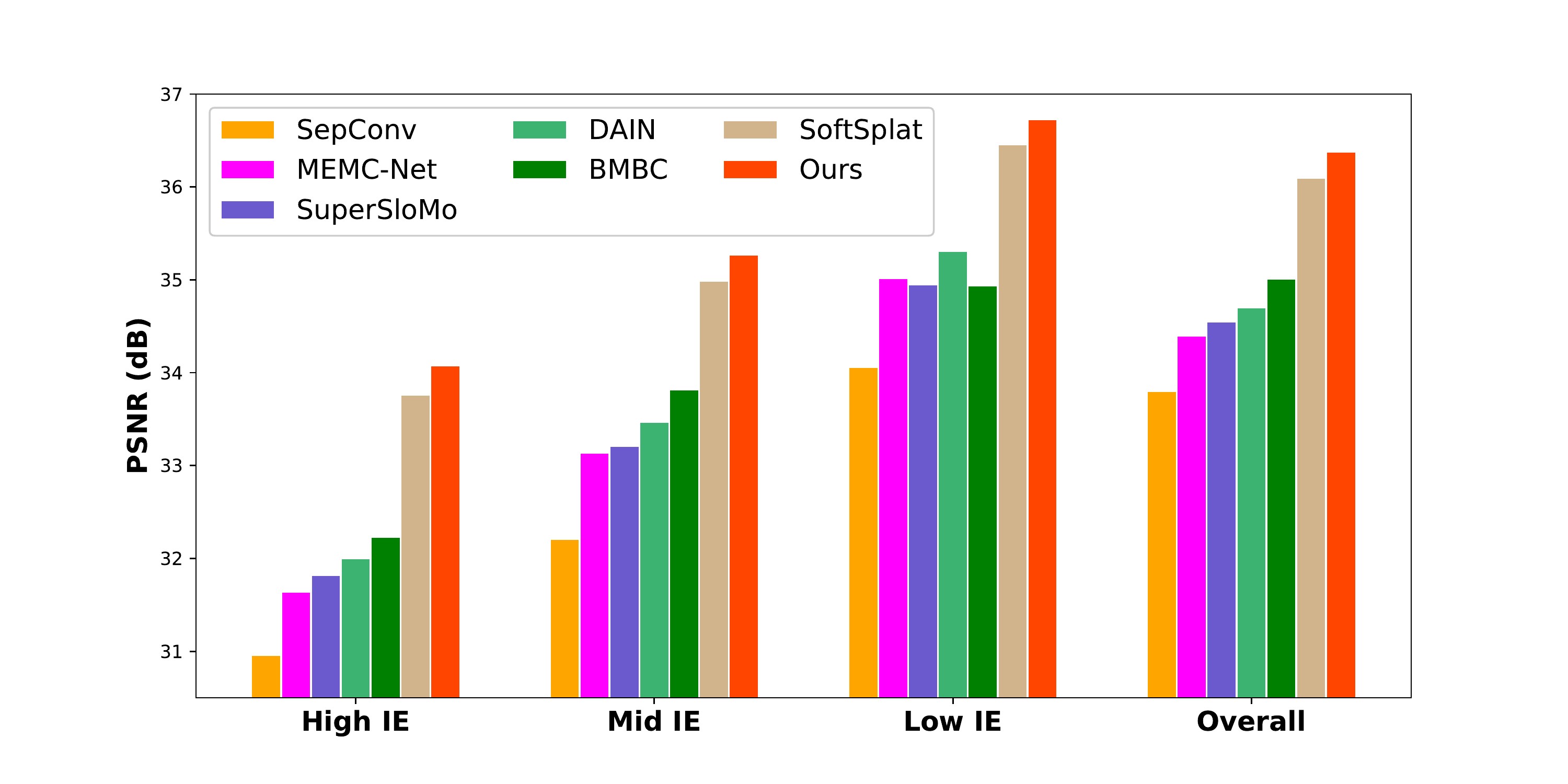}
    \end{subfigure}
    \caption{Regional comparison on Vimeo90K.}
    \label{fig: regional comparison}
\end{figure}

\noindent\textbf{Regional comparison:} A clear way to investigate the impact of our spatially adaptive frame interpolation is to compare the performance in separate regions. According to Fig.~\ref{fig: regional comparison}, for all methods, the highest and lowest PSNR are observed for low IE and high IE regions, respectively. Thus, it indicates the effectiveness of our error prediction. %\st{On the other hand, the highest level of improvement for our method is observed in high IE region. It justifies our design principle, which emphasizes more on the regions with higher space for improvement to boost the overall performance.} 

Fig.~\ref{fig: regional comparison} also reveals that our proposed method has improved the PSNR by a large margin in all three IE regions. By passing the refined region and deep features to the next stage, each stage is able to focus more on the target region and provide auxiliary improvement for other regions. Therefore, a significant improvement in low and mid IE regions is also observed. %\st{More analysis of regional evaluation can be found in the supplementary material.}

% reginal analysis

\begin{figure*}[t]
    \centering    % First row
    \small
    \begin{subfigure}{0.105\linewidth}
    \includegraphics[height=0.75\linewidth, width =\linewidth]{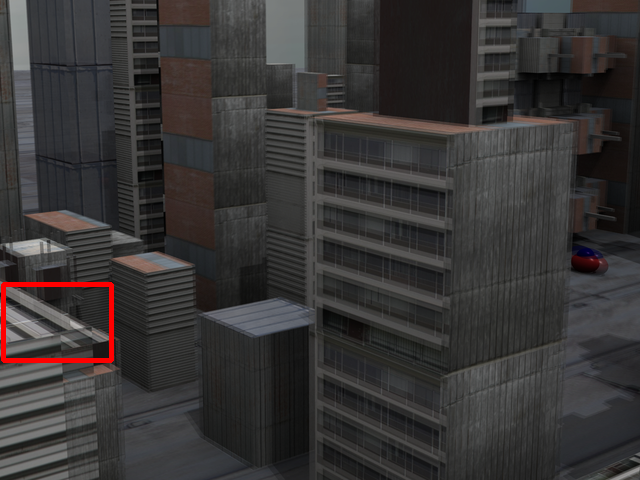}
    \end{subfigure}
    \begin{subfigure}{0.105\linewidth}
    \includegraphics[height=0.75\linewidth, width =\linewidth]{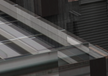}
    \end{subfigure}
    \begin{subfigure}{0.105\linewidth}
    \includegraphics[height=0.75\linewidth, width =\linewidth]{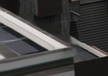}
    \end{subfigure}
    \begin{subfigure}{0.105\linewidth}
    \includegraphics[height=0.75\linewidth, width =\linewidth]{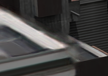}
    \end{subfigure}
    \begin{subfigure}{0.105\linewidth}
    \includegraphics[height=0.75\linewidth, width =\linewidth]{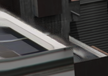}
    \end{subfigure}
    \begin{subfigure}{0.105\linewidth}
    \includegraphics[height=0.75\linewidth, width =\linewidth]{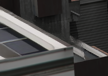}
    \end{subfigure}
    \begin{subfigure}{0.105\linewidth}
    \includegraphics[height=0.75\linewidth, width =\linewidth]{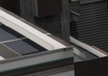}
    \end{subfigure}
    \begin{subfigure}{0.105\linewidth}
    \includegraphics[height=0.75\linewidth, width =\linewidth]{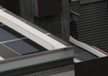}
    \end{subfigure}
    \begin{subfigure}{0.105\linewidth}
    \includegraphics[height=0.75\linewidth, width =\linewidth]{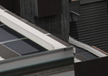}
    \end{subfigure}
    
    \vspace{0.05cm}
    \begin{subfigure}{0.105\linewidth}% second tow
    \includegraphics[height=0.92\linewidth, width =\linewidth]{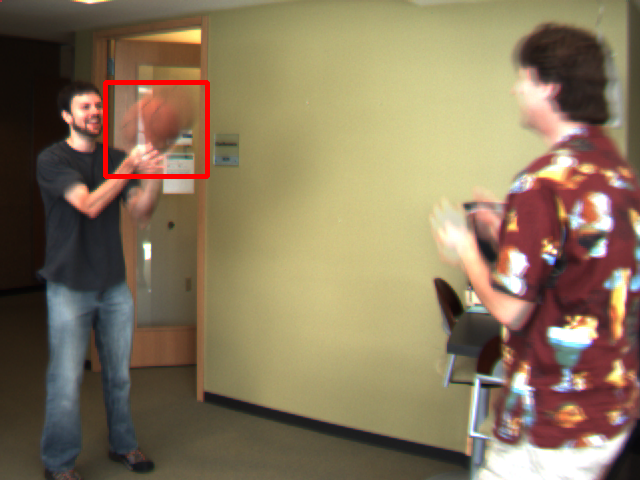}
    \subcaption*{}
    \end{subfigure}
    \begin{subfigure}{0.105\linewidth}
    \includegraphics[width =\linewidth]{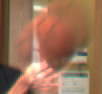}
    \subcaption*{Inputs}
    \end{subfigure}
    \begin{subfigure}{0.105\linewidth}
    \includegraphics[width =\linewidth]{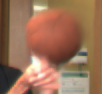}
    \subcaption*{SuperSloMo}
    \end{subfigure}
    \begin{subfigure}{0.105\linewidth}
    \includegraphics[width =\linewidth]{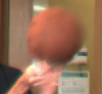}
    \subcaption*{MEMC-Net}
    \end{subfigure}
    \begin{subfigure}{0.105\linewidth}
    \includegraphics[width =\linewidth]{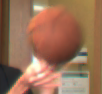}
    \subcaption*{DAIN}
    \end{subfigure}
    \begin{subfigure}{0.105\linewidth}
    \includegraphics[width =\linewidth]{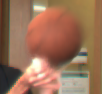}
    \subcaption*{BMBC}
    \end{subfigure}
    \begin{subfigure}{0.105\linewidth}
    \includegraphics[width =\linewidth]{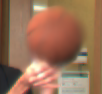}
    \subcaption*{SoftSplat}
    \end{subfigure}
    \begin{subfigure}{0.105\linewidth}
    \includegraphics[width =\linewidth]{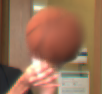}
    \subcaption*{Ours($\mathcal{L}_{ecc}$)}
    \end{subfigure}
    \begin{subfigure}{0.105\linewidth}
    \includegraphics[width =\linewidth]{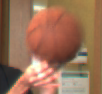}
    \subcaption*{Ours($\mathcal{L}_{ecp}$)}
    \end{subfigure}

    % UCF #1
    \begin{subfigure}{0.105\linewidth}
    \includegraphics[height=0.75\linewidth, width =\linewidth]{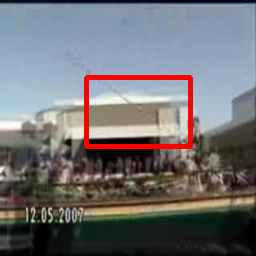}
    \end{subfigure}
    \begin{subfigure}{0.105\linewidth}
    \includegraphics[height=0.75\linewidth, width =\linewidth]{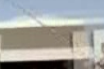}
    \end{subfigure}
    % \begin{subfigure}{0.105\linewidth}
    % \includegraphics[height=0.75\linewidth, width =\linewidth]{figs/plots/UCF_2561/1_frame_2561_super.png}
    % \end{subfigure}
    \begin{subfigure}{0.105\linewidth}
    \includegraphics[height=0.75\linewidth, width =\linewidth]{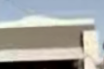}
    \end{subfigure}
    \begin{subfigure}{0.105\linewidth}
    \includegraphics[height=0.75\linewidth, width =\linewidth]{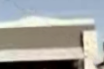}
    \end{subfigure}
    \begin{subfigure}{0.105\linewidth}
    \includegraphics[height=0.75\linewidth, width =\linewidth]{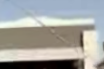}
    \end{subfigure}
    \begin{subfigure}{0.105\linewidth}
    \includegraphics[height=0.75\linewidth, width =\linewidth]{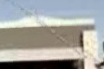}
    \end{subfigure}
    \begin{subfigure}{0.105\linewidth}
    \includegraphics[height=0.75\linewidth, width =\linewidth]{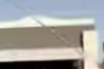}
    \end{subfigure}
    \begin{subfigure}{0.105\linewidth}
    \includegraphics[height=0.75\linewidth, width =\linewidth]{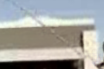}
    \end{subfigure}
    \begin{subfigure}{0.105\linewidth}
    \includegraphics[height=0.75\linewidth, width =\linewidth]{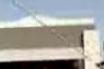}
    \end{subfigure}
    
    % UCF #2
    \begin{subfigure}{0.105\linewidth}
    \includegraphics[height=0.75\linewidth, width =\linewidth]{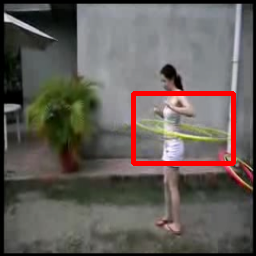}
    \subcaption*{}
    \end{subfigure}
    \begin{subfigure}{0.105\linewidth}
    \includegraphics[height=0.75\linewidth, width =\linewidth]{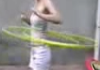}
    \subcaption*{Inputs}
    \end{subfigure}
    % \begin{subfigure}{0.105\linewidth}
    % \includegraphics[height=0.75\linewidth, width =\linewidth]{figs/plots/UCF_1611/1_frame_1611_super.png}
    % \subcaption*{SuperSloMo}
    % \end{subfigure}
    \begin{subfigure}{0.105\linewidth}
    \includegraphics[height=0.75\linewidth, width =\linewidth]{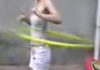}
    \subcaption*{MEMC-Net}
    \end{subfigure}
    \begin{subfigure}{0.105\linewidth}
    \includegraphics[height=0.75\linewidth, width =\linewidth]{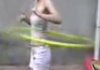}
    \subcaption*{DAIN}
    \end{subfigure}
    \begin{subfigure}{0.105\linewidth}
    \includegraphics[height=0.75\linewidth, width =\linewidth]{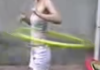}
    \subcaption*{BMBC}
    \end{subfigure}
    \begin{subfigure}{0.105\linewidth}
    \includegraphics[height=0.75\linewidth, width =\linewidth]{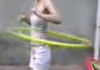}
    \subcaption*{SoftSplat}
    \end{subfigure}
    \begin{subfigure}{0.105\linewidth}
    \includegraphics[height=0.75\linewidth, width =\linewidth]{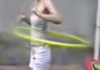}
    \subcaption*{Ours($\mathcal{L}_{ecc}$)}
    \end{subfigure}
    \begin{subfigure}{0.105\linewidth}
    \includegraphics[height=0.75\linewidth, width =\linewidth]{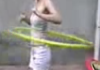}
    \subcaption*{Ours($\mathcal{L}_{ecp}$)}
    \end{subfigure}
    \begin{subfigure}{0.105\linewidth}
    \includegraphics[height=0.75\linewidth, width =\linewidth]{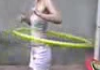}
    \subcaption*{GT}
    \end{subfigure}
    
    \vspace{0.05cm}
    \begin{subfigure}{0.17\linewidth}
    \includegraphics[width =\linewidth]{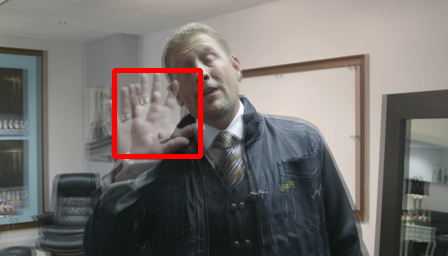}
    \end{subfigure}
    \begin{subfigure}{0.097\linewidth}
    \includegraphics[width =\linewidth]{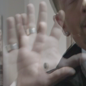}
    \end{subfigure}
    % \begin{subfigure}{0.097\linewidth}
    % \includegraphics[width =\linewidth]{figs/plots/Vimeo_00058_0658/1_sep.png}
    % \end{subfigure}
    \begin{subfigure}{0.097\linewidth}
    \includegraphics[width =\linewidth]{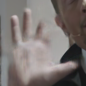}
    \end{subfigure}
    \begin{subfigure}{0.097\linewidth}
    \includegraphics[width =\linewidth]{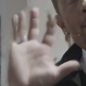}
    \end{subfigure}
    \begin{subfigure}{0.097\linewidth}
    \includegraphics[width =\linewidth]{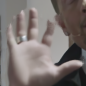}
    \end{subfigure}
    \begin{subfigure}{0.097\linewidth}
    \includegraphics[width =\linewidth]{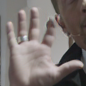}
    \end{subfigure}
    \begin{subfigure}{0.097\linewidth}
    \includegraphics[width =\linewidth]{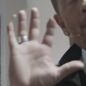}
    \end{subfigure}
    \begin{subfigure}{0.097\linewidth}
    \includegraphics[width =\linewidth]{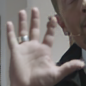}
    \end{subfigure}
    \begin{subfigure}{0.097\linewidth}
    \includegraphics[width =\linewidth]{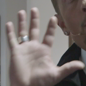}
    \end{subfigure}
    
    \vspace{0.05cm}
    \begin{subfigure}{0.17\linewidth}
    \includegraphics[width =\linewidth]{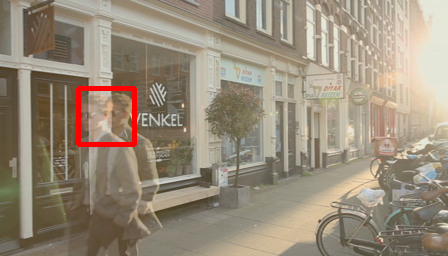}
    \subcaption*{}
    \end{subfigure}
    \begin{subfigure}{0.097\linewidth}
    \includegraphics[width =\linewidth]{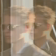}
    \subcaption*{Inputs}
    \end{subfigure}
    % \begin{subfigure}{0.097\linewidth}
    % \includegraphics[width =\linewidth]{figs/plots/Vimeo_00078_0642/1_sep.png}
    % \subcaption*{SepConv}
    % \end{subfigure}
    \begin{subfigure}{0.097\linewidth}
    \includegraphics[width =\linewidth]{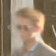}
    \subcaption*{MEMC-Net}
    \end{subfigure}
    \begin{subfigure}{0.097\linewidth}
    \includegraphics[width =\linewidth]{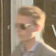}
    \subcaption*{DAIN}
    \end{subfigure}
    \begin{subfigure}{0.097\linewidth}
    \includegraphics[width =\linewidth]{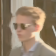}
    \subcaption*{BMBC}
    \end{subfigure}
    \begin{subfigure}{0.097\linewidth}
    \includegraphics[width =\linewidth]{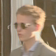}
    \subcaption*{SoftSplat}
    \end{subfigure}
    \begin{subfigure}{0.097\linewidth}
    \includegraphics[width =\linewidth]{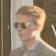}
    \subcaption*{Ours($\mathcal{L}_{ecc}$)}
    \end{subfigure}
    \begin{subfigure}{0.097\linewidth}
    \includegraphics[width =\linewidth]{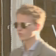}
    \subcaption*{Ours($\mathcal{L}_{ecp}$)}
    \end{subfigure}
    \begin{subfigure}{0.097\linewidth}
    \includegraphics[width =\linewidth]{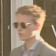}
    \subcaption*{GT}
    \end{subfigure}
    
    \caption{Qualitative comparison on Middlebury Evaluation~\cite{baker2011database}(row 1,2), UCF101~\cite{liu2017video}(row 3,4), Vimeo90K~\cite{xue2019video}(row 5,6).}
    \label{fig:qualitative}
\end{figure*}

\noindent\textbf{Qualitative analysis:} Fig.~\ref{fig:qualitative} shows the qualitative results from the Middlebury Evaluation, UCF101, and Vimeo90K datasets. The proposed method can better handle challenging scenarios. Particularly, as it can be seen in Fig.~\ref{fig:qualitative}, with $\mathcal{L}_{ecp}$, our method preserves high-frequency components and generates more visually appealing interpolated results, making it effective in practice. We also provide a demo video in the supplement to exam the visual quality and temporal consistency of our method.

\subsection{Ablation studies}

\noindent\textbf{Network structures: }
In this sub-section, we investigate each component of the network, including 1) pixel level refinement; 2) network size; 3) spatial ensembles; 4) feature inheritance. The related results are reported in Table~\ref{tab: ablation}. According to the first section of Table~\ref{tab: ablation}, comparing 1-stage with 1-stage-refine, the pixel refinement is greatly effective as it corrects the potential errors triggered during frame synthesis. However, naively increasing the network size in the second section of Table~\ref{tab: ablation} (comparing 1 stage to 3 stages network as in Fig.~\ref{fig: architeture}a) does not give obvious improvement and suffers from performance saturation. To show the effectiveness of spatial adaptive processing, we train the networks in Fig.~\ref{fig: architeture}b and c and denote them as \textit{Parallel} and \textit{Proposed} (second section of Table~\ref{tab: ablation}). For \textit{Parallel}, the middle frame is obtained by adding the processed regions and further processed by a 2-layer network to deal with the boundary issues. As we can see, partitioning the regions according to error level pushes the performance limitation. And the proposed ensemble approach is more optimized and further improves the results. Finally, feature inheritance brings additional improvement, as the subsequent stages can avoid computing the redundant information from previous stages.

% Ablation structure
\begin{table}[t]
% \scriptsize
\small
% \footnotesize
\centering
\setlength{\tabcolsep}{1.3pt} % Default value: 6pt

\begin{tabular}{lcccccc}
\toprule
\multirow{2}{*}{Methods} & \multicolumn{3}{c}{Vimeo90K~\cite{xue2019video}} & \multicolumn{3}{c}{UCF101~\cite{liu2017video}}\\
 \cmidrule(r){2-4} \cmidrule(r){5-7}
 & PSNR$\uparrow$& SSIM$\uparrow$ & LPIPS$\downarrow$ & PSNR$\uparrow$ & SSIM$\uparrow$ & LPIPS$\downarrow$ \\
\midrule
1-stage &35.30 &0.965 &0.024 &35.00 &0.947 &0.034 \\
1-stage-refine &35.48 &0.967 &0.023 &35.07 &0.948 &0.034 \\
\midrule
1-stage-refine &35.48 &0.967 &0.023 &35.07 &0.948 &0.034 \\
3-stage-refine  &35.58 &0.967 &0.023 &35.09 & 0.948&0.034 \\
Parallel & 35.86 & 0.969 & 0.022 & 35.20 & 0.949 & 0.033 \\ 
Proposed  & 36.11 & 0.970 & 0.022 & 35.31 & 0.951 & 0.032 \\ 
Proposed-feat. & 36.38 & 0.972 & 0.020 & 35.41 & 0.952 & 0.031 \\
\midrule
Metric-$\mathcal{E}_{pc}$ &36.04 &0.970 &0.023 &35.30 &0.950 &0.033 \\
Metric-$\mathcal{E}_{mv}$ &36.21 &0.971 &0.021 &35.38 &0.951 &0.032 \\
Metric-$\mathcal{E}_{ms}$ &36.30 &0.971 &0.021 &35.39 &0.951 &0.032 \\
Metric-$\mathcal{E}_{tot}$ &36.38 & 0.972 & 0.020 & 35.41 & 0.952 & 0.031 \\
\midrule
Ours-Compact & 35.01 & 0.963& 0.023& 35.13& 0.950& 0.034\\
Ours-SPyNet & 36.18 & 0.970 & 0.022& 35.36 & 0.951& 0.032\\
Ours-PWC-Net & 36.27 & 0.971 & 0.021& 35.39 & 0.952 & 0.032\\
Ours-RAFT & 36.38 & 0.972 & 0.020 & 35.41 & 0.952 & 0.031 \\
\bottomrule
\end{tabular}
\caption{Ablation studies on network structures, error prediction metrics and optical flow networks.
}
\label{tab: ablation}
% \vspace{-3mm}
\end{table}

\noindent\textbf{Error prediction metrics:} To investigate the impact of applying the proposed error-aware adaptive strategy. We train 4 models with each of the proposed error prediction metrics ($\mathcal{E}_{ms}, \mathcal{E}_{mv}, \mathcal{E}_{pc}$ and $\mathcal{E}_{tot}$). As reported in the third section of Table~\ref{tab: ablation}, the adaptive error-aware mechanism brings significant improvement for all metrics. Among the prediction metrics, $\mathcal{E}_{pc}$ brings less impact, as it mostly detects occlusions that normally occupy a small portion of the frames. Utilizing the OF statistics yields better results, as it directly reflects interpolation complexity across the frame. 

\begin{figure}[t]
    \centering
    \begin{subfigure}{\linewidth}
    \includegraphics[width =\linewidth]{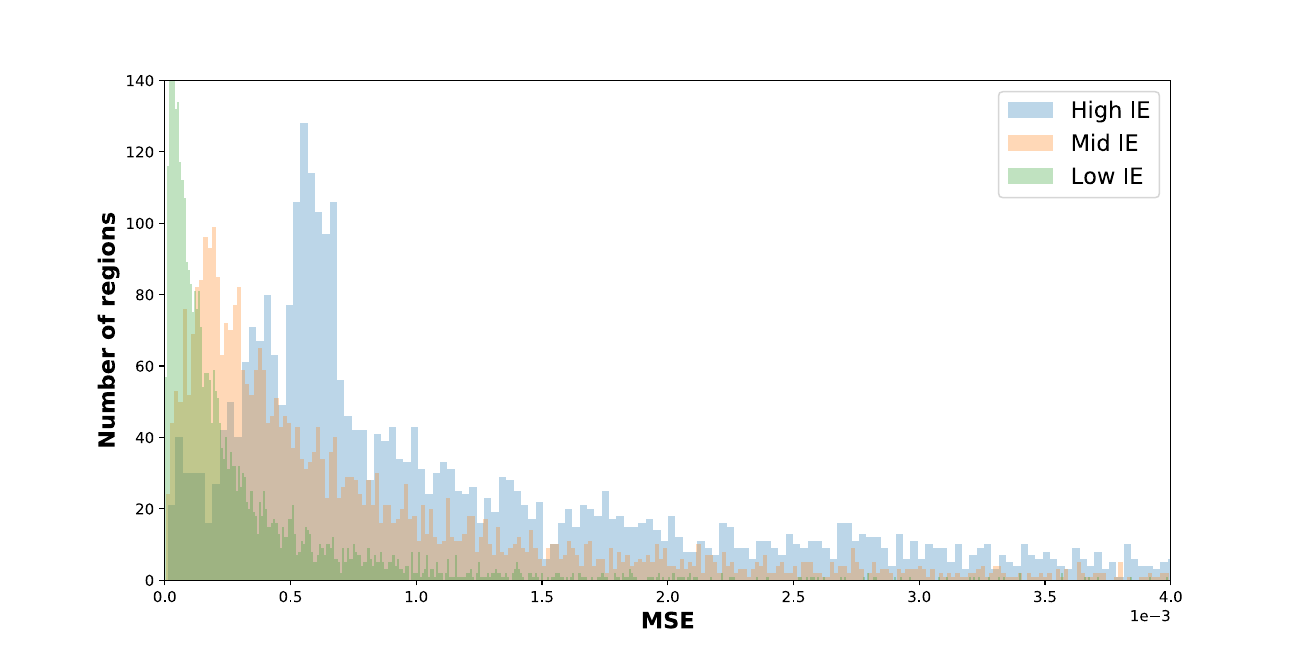}
    \end{subfigure}
    \caption{The distribution of IE measured as MSE in three regions classified by thresholding $\mathcal{E}_{tot}$.}
    \label{fig: mse}
\end{figure}

To illustrate the correlation between the proposed metrics and IE, we compare the PSNR of high IE region vs. the entire frame. We analyze the model \textit{3-stage} on the validation set. As shown in Fig.~\ref{fig: PSNR scatter}, the points under the diagonal line are the samples where the PSNR of the selected high IE region is less than the whole frame. In fact, going more down the diagonal means the selected high IE region is experiencing higher IE. As we can see, more samples are dispersedly distributed below the diagonal when the error is predicted by $\mathcal{E}_{ms}$. Therefore, it performs better, as reported in Table~\ref{tab: ablation}. Incorporating all metrics ($\mathcal{E}_{tot}$) brings additional improvement as they compensate each other in finding more regions with higher error. Fig.~\ref{fig: mse} shows the MSE distributions of three regions. The three distinct distributions illustrate the error-based segmentation is meaningful.

% MSE distribution
% Efficiency    
\begin{figure}
    \centering
    \begin{subfigure}{\linewidth}
    \includegraphics[width =0.9\linewidth]{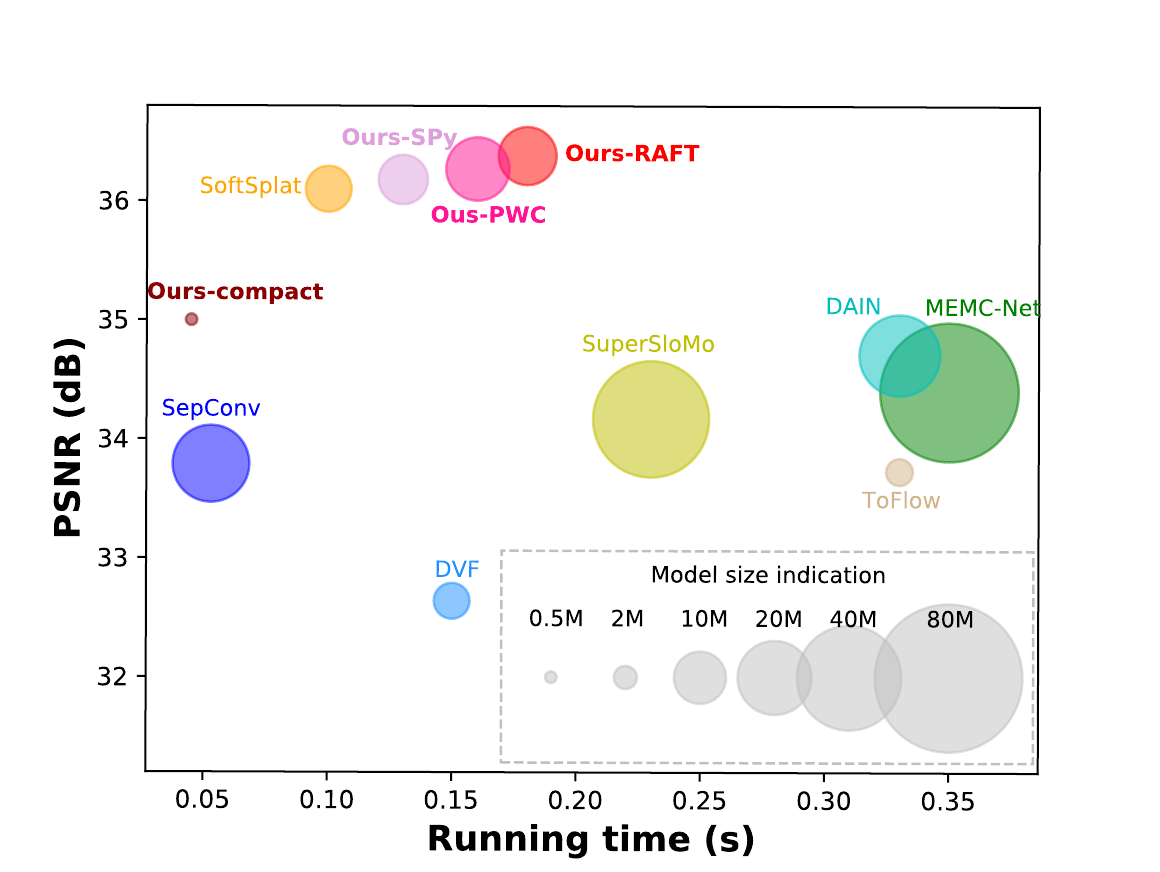}
    \end{subfigure}
    \caption{Trade-off comparison on PSNR, speed and model size. The model size is visualized by the circle radius.
    }
    \label{fig: efficiency}
\end{figure}

\noindent\textbf{Optical flow networks:}
We further study the impacts from various OF estimation networks, including PWC-Net~\cite{sun2018pwc}, SPyNet~\cite{ranjan2017optical} and RAFT~\cite{teed2020raft}. As reported in the last section of Table~\ref{tab: ablation}, they perform comparably well due to the error-adaptive mechanism. It is noted that the network trained with SPyNet~\cite{ranjan2017optical} performs favorably against SoftSplat~\cite{niklaus2020softmax} where a better flow network, PWC-Net, is used.

\noindent\textbf{Compact model:}
The error prediction and spatially adaptive processing reach a more optimized solution; it allows us to design a compact model for low-power devices. We first integrate the frame synthesis module into SPyNet such that it directly computes $f_{0 \rightarrow t}$, $f_{1 \rightarrow t}$ and visibility map to generate the initial middle frame. We further reduce the layers and channels so that it contains only 0.59 million parameters. As reported in Table~\ref{tab: compare SOTA}, the compact version achieves similar quantitative results compared to DAIN~\cite{DAIN}, but it is 40 times smaller in model size. We also record inference time on interpolating a $480 \times 640$ image. As shown in Fig.~\ref{fig: efficiency}, our compact model achieves a better trade-off in speed-quality-size.

\begin{figure}[t]
    \centering
    
    %%%%% First example %%%%%
    \begin{subfigure}{0.19\linewidth} 
    \includegraphics[width =\linewidth]{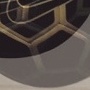}
    \end{subfigure}
    \begin{subfigure}{0.19\linewidth} 
    \includegraphics[width =\linewidth]{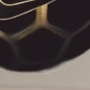}
    \end{subfigure}
    \begin{subfigure}{0.19\linewidth} 
    \includegraphics[width =\linewidth]{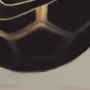}
    \end{subfigure}
    \begin{subfigure}{0.19\linewidth} 
    \includegraphics[width =\linewidth]{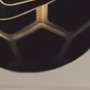}
    \end{subfigure}
    \begin{subfigure}{0.19\linewidth} 
    \includegraphics[width =\linewidth]{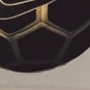}
    \end{subfigure}
    %%%%% second example %%%%% 
    \begin{subfigure}[t]{0.19\linewidth} 
    \includegraphics[width =\linewidth]{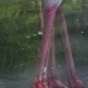}
    \caption{\footnotesize Input}
    \end{subfigure}
    \begin{subfigure}[t]{0.19\linewidth} 
    \includegraphics[width =\linewidth]{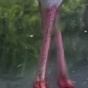}
    \caption{\footnotesize No adap.}
    \end{subfigure}
    \begin{subfigure}[t]{0.19\linewidth}
    \includegraphics[width =\linewidth]{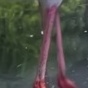}
    \caption{\footnotesize $\mathcal{L}_{ecc}^m$}
    \end{subfigure}
    \begin{subfigure}[t]{0.19\linewidth} 
    \includegraphics[width =\linewidth]{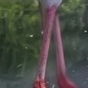}
    \caption{\footnotesize $\mathcal{L}_{ecc}$}
    \end{subfigure}
    \begin{subfigure}[t]{0.19\linewidth} 
    \includegraphics[width =\linewidth]{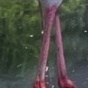}
    \caption{\footnotesize $\mathcal{L}_{ecp}$}
    \end{subfigure}
    \caption{Visual comparison on different loss functions.}
    \label{fig:ablation loss ims}
\end{figure}

\noindent\textbf{Loss functions:} The error-controlled loss function plays an important role in adaptive processing, as it does not attenuate the optimization in higher IE regions. To demonstrate its effectiveness, we train another model with a variant of $\mathcal{L}_{ecc}$, where the masks in each stage are \textit{merged} by union operator. We denote it as $\mathcal{L}_{ecc}^m$. As reported in Table~\ref{tab: ablation loss}, the performance of the model trained with $\mathcal{L}_{ecc}^m$ drops as expected. While keeping the regions separate when calculating the loss yields better results as it facilitates better regional gradient computation. As also reported in Table~\ref{tab: ablation loss}, the model trained by the perceptual loss $\mathcal{L}_{ecp}$ performs best in terms of LPIPS. Better fine details are also restored by $\mathcal{L}_{ecp}$, as shown in Fig.~\ref{fig:ablation loss ims}.

% Ablation loss function
\begin{table}[t]
% \scriptsize
\small
% \footnotesize
\centering
\setlength{\tabcolsep}{2pt} % Default value: 6pt

\begin{tabular}{lcccccc}
\toprule
\multirow{2}{*}{Methods} & \multicolumn{3}{c}{Vimeo90K~\cite{xue2019video}} & \multicolumn{3}{c}{DAVIS~\cite{pont20172017}}\\
 \cmidrule(r){2-4} \cmidrule(r){5-7}
 & PSNR$\uparrow$ & SSIM$\uparrow$ & LPIPS$\downarrow$ & PSNR$\uparrow$ & SSIM$\uparrow$ & LPIPS$\downarrow$ \\
 
\midrule
No adap.-$\mathcal{L}_1$ & 35.58 &0.967 &0.023 & 26.91& 0.799& 0.125\\
Ours-$\mathcal{L}_{ecc}^{m}$ &36.15 &0.970 &0.021 &27.29 &0.808 &0.117\\
Ours-$\mathcal{L}_{ecc}$ &\textbf{36.38} &\textbf{0.972} &0.020 &\textbf{27.55} &\textbf{0.815} &0.111\\
Ours-$\mathcal{L}_{ecp}$ &35.95 &0.967 &\textbf{0.012} &26.98 &0.792 &\textbf{0.080}\\
\bottomrule
\end{tabular}
\caption{Ablation studies on error controlled loss.
}
\label{tab: ablation loss}
% \vspace{-3mm}
\end{table}

% ablation loss function

\section{Conclusions}
This paper proposed a novel, spatially adaptive, error-aware, network ensemble for video frame interpolation. The developed solution was based on a key observation, namely the spatial variance of the motion complexity and the non-uniformity of the resulting interpolation error. By investigating the behavior of optical flow, we are able to partition the frames according to IE error levels, and introduce an ensemble to process segments based on predicted regional errors. An efficient and cost effective solution was obtained. The comprehensive experimentation conducted using benchmark datasets provided substantial evidence of the solution's utility. 

% Future works will include non-heuristic thresholding and further analysis on the utility of the proposed here error metrics.

{\small
\bibliographystyle{ieee_fullname}
\bibliography{egbib}

\begin{thebibliography}{10}\itemsep=-1pt

\bibitem{ensemble}
Stratos~Idreos Abdul~Wasay.
\newblock More or less: When and how to build convolutional neural network
  ensembles.
\newblock In {\em International Conference on Learning Representations}, 2021.

\bibitem{ashukha2020pitfalls}
Arsenii Ashukha, Alexander Lyzhov, Dmitry Molchanov, and Dmitry Vetrov.
\newblock Pitfalls of in-domain uncertainty estimation and ensembling in deep
  learning.
\newblock In {\em International Conference on Learning Representations}, 2020.

\bibitem{baker2011database}
Simon Baker, Daniel Scharstein, JP Lewis, Stefan Roth, Michael~J Black, and
  Richard Szeliski.
\newblock A database and evaluation methodology for optical flow.
\newblock {\em International Journal of Computer Vision}, 92(1):1--31, 2011.

\bibitem{DAIN}
Wenbo Bao, Wei-Sheng Lai, Chao Ma, Xiaoyun Zhang, Zhiyong Gao, and Ming-Hsuan
  Yang.
\newblock Depth-aware video frame interpolation.
\newblock In {\em IEEE Conferene on Computer Vision and Pattern Recognition},
  2019.

\bibitem{bao2018memc}
Wenbo Bao, Wei-Sheng Lai, Xiaoyun Zhang, Zhiyong Gao, and Ming-Hsuan Yang.
\newblock Memc-net: Motion estimation and motion compensation driven neural
  network for video interpolation and enhancement.
\newblock {\em IEEE Transactions on Pattern Analysis and Machine Intelligence},
  2019.

\bibitem{chi2020all}
Zhixiang Chi, Rasoul~Mohammadi Nasiri, Zheng Liu, Juwei Lu, Jin Tang, and
  Konstantinos~N Plataniotis.
\newblock All at once: Temporally adaptive multi-frame interpolation with
  advanced motion modeling.
\newblock In {\em European Confererence on Computer Vison}, 2020.

\bibitem{dufaux1996segmentation}
Fr{\'e}d{\'e}ric Dufaux and Fabrice Moscheni.
\newblock Segmentation-based motion estimation for second generation video
  coding techniques.
\newblock In {\em Video Coding}, pages 219--263. Springer, 1996.

\bibitem{hinton2015distilling}
Geoffrey Hinton, Oriol Vinyals, and Jeff Dean.
\newblock Distilling the knowledge in a neural network.
\newblock In {\em Advances in Neural Information Processing Systems Deep
  Learning Workshop}, 2014.

\bibitem{jiang2018super}
Huaizu Jiang, Deqing Sun, Varun Jampani, Ming-Hsuan Yang, Erik Learned-Miller,
  and Jan Kautz.
\newblock Super slomo: High quality estimation of multiple intermediate frames
  for video interpolation.
\newblock In {\em IEEE Conference on Computer Vision and Pattern Recognition},
  2018.

\bibitem{katsenou2018exploring}
Angeliki~V Katsenou, Alex Mackin, D Ma, F Zhang, and DR Bull.
\newblock Exploring the challenges of higher frame rates: from quality
  assessment to frame rate selection.
\newblock In {\em IEEE COMSOC MMTC Communications-Frontiers (E-Letter)}, 2018.

\bibitem{kingma2014adam}
Diederik~P Kingma and Jimmy Ba.
\newblock Adam: A method for stochastic optimization.
\newblock In {\em International Conference on Learning Representations}, 2015.

\bibitem{kunt1985second}
Murat Kunt, Athanassios Ikonomopoulos, and Michel Kocher.
\newblock Second-generation image-coding techniques.
\newblock {\em Proceedings of the IEEE}, 73(4):549--574, 1985.

\bibitem{lakshminarayanan2016simple}
Balaji Lakshminarayanan, Alexander Pritzel, and Charles Blundell.
\newblock Simple and scalable predictive uncertainty estimation using deep
  ensembles.
\newblock In {\em Advances in Neural Information Processing Systems}, 2017.

\bibitem{lazebnik2006beyond}
Svetlana Lazebnik, Cordelia Schmid, and Jean Ponce.
\newblock Beyond bags of features: Spatial pyramid matching for recognizing
  natural scene categories.
\newblock In {\em IEEE Computer Society Conference on Computer Vision and
  Pattern Recognition}, 2006.

\bibitem{li2019lap}
Yunan Li, Qiguang Miao, Wanli Ouyang, Zhenxin Ma, Huijuan Fang, Chao Dong, and
  Yining Quan.
\newblock Lap-net: Level-aware progressive network for image dehazing.
\newblock In {\em IEEE International Conference on Computer Vision}, 2019.

\bibitem{liu2019deep}
Yu-Lun Liu, Yi-Tung Liao, Yen-Yu Lin, and Yung-Yu Chuang.
\newblock Deep video frame interpolation using cyclic frame generation.
\newblock In {\em AAAI Conference on Artificial Intelligence}, 2019.

\bibitem{liu2017video}
Ziwei Liu, Raymond~A Yeh, Xiaoou Tang, Yiming Liu, and Aseem Agarwala.
\newblock Video frame synthesis using deep voxel flow.
\newblock In {\em IEEE International Conference on Computer Vision}, 2017.

\bibitem{meyer2018phasenet}
Simone Meyer, Abdelaziz Djelouah, Brian McWilliams, Alexander Sorkine-Hornung,
  Markus Gross, and Christopher Schroers.
\newblock Phasenet for video frame interpolation.
\newblock In {\em IEEE Conference on Computer Vision and Pattern Recognition},
  2018.

\bibitem{nah2017deep}
Seungjun Nah, Tae Hyun~Kim, and Kyoung Mu~Lee.
\newblock Deep multi-scale convolutional neural network for dynamic scene
  deblurring.
\newblock In {\em IEEE Conference on Computer Vision and Pattern Recognition},
  2017.

\bibitem{nasiri2018temporal}
Rasoul~Mohammadi Nasiri, Zhengfang Duanmu, and Zhou Wang.
\newblock Temporal motion smoothness and the impact of frame rate variation on
  video quality.
\newblock In {\em IEEE International Conference on Image Processing}, 2018.

\bibitem{nasiri2015perceptual}
Rasoul~Mohammadi Nasiri, Jiheng Wang, Abdul Rehman, Shiqi Wang, and Zhou Wang.
\newblock Perceptual quality assessment of high frame rate video.
\newblock In {\em International Workshop on Multimedia Signal Processing},
  2015.

\bibitem{niklaus2018context}
Simon Niklaus and Feng Liu.
\newblock Context-aware synthesis for video frame interpolation.
\newblock In {\em IEEE Conference on Computer Vision and Pattern Recognition},
  2018.

\bibitem{niklaus2020softmax}
Simon Niklaus and Feng Liu.
\newblock Softmax splatting for video frame interpolation.
\newblock In {\em IEEE Conference on Computer Vision and Pattern Recognition},
  2020.

\bibitem{niklaus2017video_1}
Simon Niklaus, Long Mai, and Feng Liu.
\newblock Video frame interpolation via adaptive convolution.
\newblock In {\em IEEE Conference on Computer Vision and Pattern Recognition},
  2017.

\bibitem{niklaus2017video}
Simon Niklaus, Long Mai, and Feng Liu.
\newblock Video frame interpolation via adaptive separable convolution.
\newblock In {\em IEEE International Conference on Computer Vision}, 2017.

\bibitem{park2020bmbc}
Junheum Park, Keunsoo Ko, Chul Lee, and Chang-Su Kim.
\newblock Bmbc: Bilateral motion estimation with bilateral cost volume for
  video interpolation.
\newblock In {\em European Confererence on Computer Vison}, 2020.

\bibitem{pont20172017}
Jordi Pont-Tuset, Federico Perazzi, Sergi Caelles, Pablo Arbel{\'a}ez, Alex
  Sorkine-Hornung, and Luc Van~Gool.
\newblock The 2017 davis challenge on video object segmentation.
\newblock {\em arXiv preprint arXiv:1704.00675}, 2017.

\bibitem{ranjan2017optical}
Anurag Ranjan and Michael~J Black.
\newblock Optical flow estimation using a spatial pyramid network.
\newblock In {\em IEEE Conference on Computer Vision and Pattern Recognition},
  2017.

\bibitem{ronneberger2015u}
Olaf Ronneberger, Philipp Fischer, and Thomas Brox.
\newblock U-net: Convolutional networks for biomedical image segmentation.
\newblock In {\em International Conference on Medical Image Computing and
  Computer Assisted Intervention}, 2015.

\bibitem{shen2018deep}
Ziyi Shen, Wei-Sheng Lai, Tingfa Xu, Jan Kautz, and Ming-Hsuan Yang.
\newblock Deep semantic face deblurring.
\newblock In {\em IEEE Conference on Computer Vision and Pattern Recognition},
  2018.

\bibitem{simonyan2014very}
Karen Simonyan and Andrew Zisserman.
\newblock Very deep convolutional networks for large-scale image recognition.
\newblock In {\em International Conference on Learning Representations}, 2015.

\bibitem{suin2020spatially}
Maitreya Suin, Kuldeep Purohit, and AN Rajagopalan.
\newblock Spatially-attentive patch-hierarchical network for adaptive motion
  deblurring.
\newblock In {\em IEEE Conference on Computer Vision and Pattern Recognition},
  2020.

\bibitem{sun2018pwc}
Deqing Sun, Xiaodong Yang, Ming-Yu Liu, and Jan Kautz.
\newblock Pwc-net: Cnns for optical flow using pyramid, warping, and cost
  volume.
\newblock In {\em IEEE Conference on Computer Vision and Pattern Recognition},
  2018.

\bibitem{teed2020raft}
Zachary Teed and Jia Deng.
\newblock Raft: Recurrent all-pairs field transforms for optical flow.
\newblock In {\em European Confererence on Computer Vison}, 2020.

\bibitem{waleed2018unreasonable}
Muhammad Waleed~Gondal, Bernhard Scholkopf, and Michael Hirsch.
\newblock The unreasonable effectiveness of texture transfer for single image
  super-resolution.
\newblock In {\em European Conference on Computer Vision}, 2018.

\bibitem{xu2019quadratic}
Xiangyu Xu, Li Siyao, Wenxiu Sun, Qian Yin, and Ming-Hsuan Yang.
\newblock Quadratic video interpolation.
\newblock In {\em Advances in Neural Information Processing Systems}, 2019.

\bibitem{xue2019video}
Tianfan Xue, Baian Chen, Jiajun Wu, Donglai Wei, and William~T Freeman.
\newblock Video enhancement with task-oriented flow.
\newblock {\em International Journal of Computer Vision}, 127(8):1106--1125,
  2019.

\bibitem{yuan2019zoom}
Liangzhe Yuan, Yibo Chen, Hantian Liu, Tao Kong, and Jianbo Shi.
\newblock Zoom-in-to-check: Boosting video interpolation via instance-level
  discrimination.
\newblock In {\em IEEE Conference on Computer Vision and Pattern Recognition},
  2019.

\bibitem{zhang2019deep}
Hongguang Zhang, Yuchao Dai, Hongdong Li, and Piotr Koniusz.
\newblock Deep stacked hierarchical multi-patch network for image deblurring.
\newblock In {\em IEEE Conference on Computer Vision and Pattern Recognition},
  2019.

\bibitem{zhang2018densely}
He Zhang and Vishal~M Patel.
\newblock Densely connected pyramid dehazing network.
\newblock In {\em IEEE Conference on Computer Vision and Pattern Recognition},
  2018.

\bibitem{zhang2018unreasonable}
Richard Zhang, Phillip Isola, Alexei~A Efros, Eli Shechtman, and Oliver Wang.
\newblock The unreasonable effectiveness of deep features as a perceptual
  metric.
\newblock In {\em IEEE Conference on Computer Vision and Pattern Recognition},
  2018.

\end{thebibliography}
}

\end{document}